\documentclass[letterpaper, 10 pt, conference]{ieeeconf}  

\IEEEoverridecommandlockouts                              

\usepackage{graphics} 
\usepackage{epsfig} 
\usepackage{mathptmx} 
\usepackage{times} 
\usepackage{amsmath, bm} 
\usepackage{amssymb}  
\usepackage{lipsum}
\usepackage{color}

\usepackage[style=ieee,dashed=false]{biblatex}
\usepackage{ifthen}
\makeatletter
\newcounter{IEEE@bibentries}
\renewcommand\IEEEtriggeratref[1]{%
  \renewbibmacro{finentry}{%
    \stepcounter{IEEE@bibentries}%
    \ifthenelse{\equal{\value{IEEE@bibentries}}{#1}}
    {\finentry\@IEEEtriggercmd}
    {\finentry}%
  }%
}
\makeatother

\DeclareSourcemap{
  \maps{
    \map{
      \pertype{article}
      \step[fieldset=url, null]
      \step[fieldset=doi, null]
      \step[fieldset=issn, null]
      \step[fieldset=isbn, null]
      \step[fieldset=note, null]
      \step[fieldset=editor, null]
      \step[fieldset=urldate, null]
      \step[fieldset=file, null]
    }
  }
}
\DeclareSourcemap{
  \maps{
    \map{
      \pertype{inproceedings}
      \step[fieldset=url, null]
      \step[fieldset=doi, null]
      \step[fieldset=issn, null]
      \step[fieldset=isbn, null]
      \step[fieldset=note, null]
      \step[fieldset=editor, null]
      \step[fieldset=urldate, null]
      \step[fieldset=file, null]
    }
  }
}
\DeclareSourcemap{
  \maps{
    \map{
      \pertype{incollection}
      \step[fieldset=url, null]
      \step[fieldset=doi, null]
      \step[fieldset=issn, null]
      \step[fieldset=isbn, null]
      \step[fieldset=note, null]
      \step[fieldset=editor, null]
      \step[fieldset=urldate, null]
      \step[fieldset=file, null]
    }
  }
}

\title{\LARGE \bf
A Letter on Progress Made on Husky Carbon: \\
A Legged-Aerial, Multi-modal Platform
}

\author{Adarsh Salagame$^{1 \dagger}$, Shoghair Manjikian$^{1 \dagger}$, Chenghao Wang$^{1 \dagger}$, \\Kaushik Venkatesh Krishnamurthy$^{1}$, Shreyansh Pitroda$^{1}$, Bibek Gupta$^{1}$, Tobias Jacob$^{1}$, \\ Benjamin Mottis$^{2}$, Eric Sihite$^{2}$, Milad Ramezani$^{3}$, and Alireza Ramezani$^{1}$
\thanks{$^{\dagger}$The authors have equal contributions to this work.}%
\thanks{$^{1}$The authors are with the Department of Electrical and Computer Engineering, Northeastern University, Boston, MA, USA. SiliconSynapse Laboratory. Emails: {\tt\small salagame.a, manjikian.s, wang.chengh, venkateshkrishnamu.k, pitroda.s@northeastern.edu, , bibekgupta@gmail.com, jacob.to, a.ramezani@northeastern.edu}.}%
\thanks{$^{2}$The authors are with the Department of Aerospace Engineering, California Institute of Technology, Pasadena, CA, USA. Email: {\tt\small bmottis, esihite@caltech.edu}.}%
\thanks{$^{3}$ The author is with the Robotics and Autonomous Systems Group, Data61, CSIRO, Brisbane, QLD, Australia. Email: {\tt\small milad.ramezani@data61.csiro.au}.}
}

\begin{document}

\maketitle
\thispagestyle{empty}
\pagestyle{empty}

\begin{abstract}


Animals, such as birds, widely use multi-modal locomotion by combining legged and aerial mobility with dominant inertial effects. The robotic biomimicry of this multi-modal locomotion feat can yield ultra-flexible systems in terms of their ability to negotiate their task spaces. The main objective of this paper is to discuss the challenges in achieving multi-modal locomotion, and to report our progress in developing our quadrupedal robot capable of multi-modal locomotion (legged and aerial locomotion), the \textit{Husky Carbon}. We report the mechanical and electrical components utilized in our robot, in addition to the simulation and experimentation done to achieve our goal in developing a versatile multi-modal robotic platform.

\end{abstract}

\section{INTRODUCTION}

Multi-modal locomotion in the form of combining legged, ballistic (aerial mobility with dominant inertial effects) or flight maneuvers is widely used by animals such as birds. Birds can employ their legs to walk very fast over flat or rough terrain in search of food. They can use their legs to jump to make transitions to flight mode. During flight mode, birds locomotion dynamics and tools applied to realize their mobility differs considerably. For instance, legged locomotion entails intermittent interactions and impulsive effects whereas aerial mobility involves the dexterous manipulation of the fluidic environment through smooth and continuous system-environment interactions. 

As it can be seen, birds combine two forms of mobility that dictate antagonistic requirements. They have achieved their capability through evolutionary journey and natural selection. Such selection among possible options that substantiate manipulation of various forces, including inertial, aerodynamics, or contact force contributions, is rarely considered in robot locomotion. In robot locomotion, these contributions are often isolated to remain only focused on the mode of interest.



The robotic biomimicry of these multi-modal locomotion feat can yield ultra-flexible systems in terms of their ability to negotiate their task spaces. However, the robotic biomimicry of birds' multi-modal locomotion can be a significant ordeal. The prohibitive design restrictions such as a tight power budget, limited payload, complex multi-modal actuation and perception, excessive number of active and passive joints involved in each mode, sophisticated control and environment-specific models, just to name a few, have alienated these concepts. 

The challenge is mechanisms experience conflicting force-motion behaviors in legged-aerial system. Attaining these structural behavior is not trivial. These conflicting requirements make designing these systems challenging, and may result in a design that performs both tasks poorly. Adaptive-property structures can offer new opportunities, however, their effectiveness is unexplored in legged-aerial robotics \cite{mintchev2017insect}. In adaptive structures, variable-stiffness regulation can produce rigid body hosting propellers or jointed structure substantiating legged locomotion.


\begin{figure}[t!]
    \centering
    \vspace{0.1in}
    \includegraphics[width=1.0\linewidth]{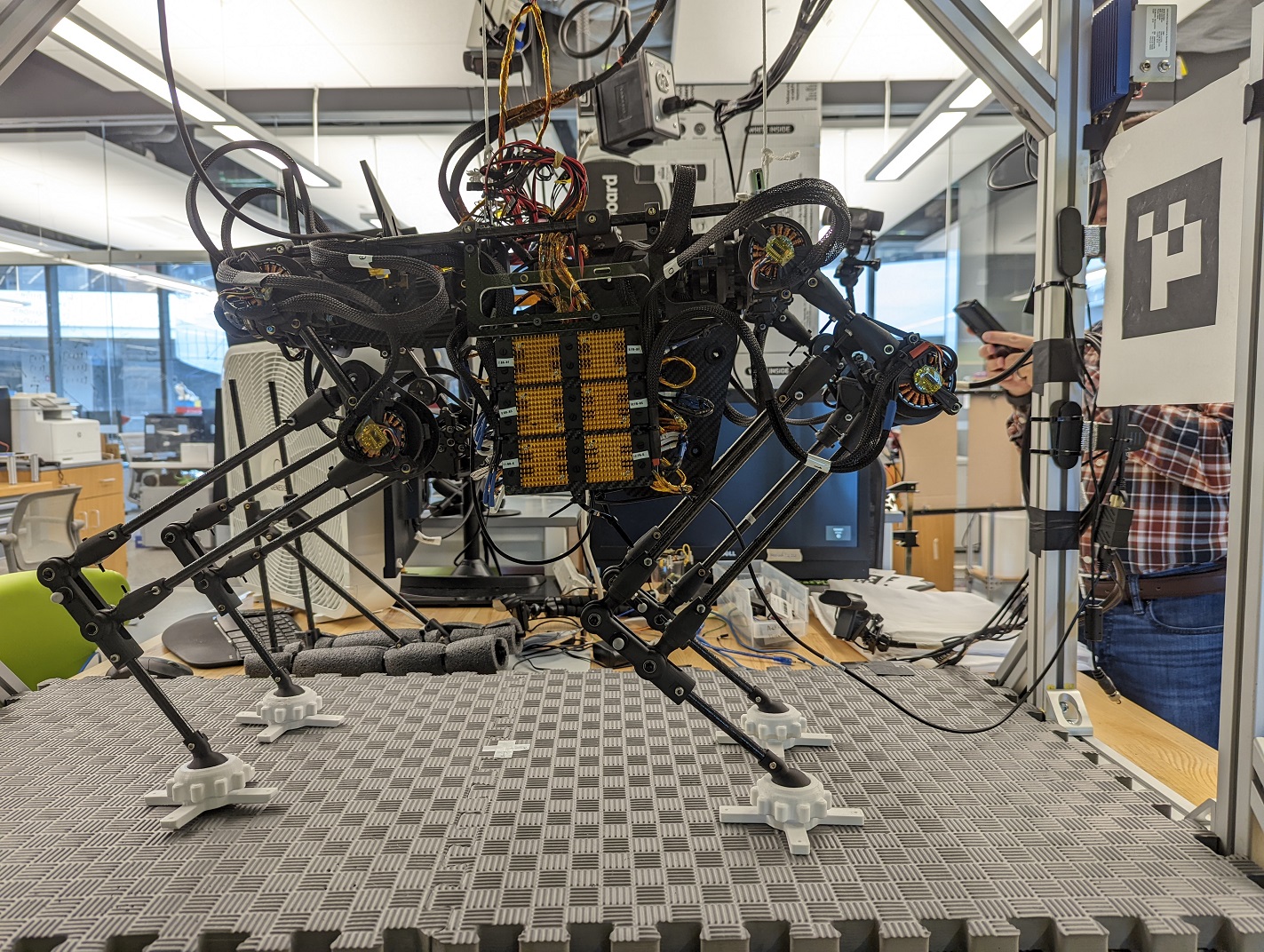}
    \caption{Husky Carbon v1.0, a quadripedal robot capable of multi-modal locomotion with ground and aerial mobility.}
    \vspace{-0.1in}
    \label{fig:cover-image}
\end{figure}

There are several works that have reported ground and aerial mobility in a single platform. However, there are a few works that have reported legged-aerial locomotion \cite{luo2015biomimetic,kim_bipedal_2021,ghassemi2016feasibility,pratt2016dynamic, shin2019development, chukewad2021robofly,roderick2021bird}. One relevant example is LEONARDO \cite{kim_bipedal_2021}, a bipedal robot. LEONARDO uses thrusters to aid in balancing, which is particularly helpful while climbing over obstacles. The addition of these thrusters also enables agile dynamic tasks such as skateboarding and tightrope walking. There are many variations to Leonardo such as BALLU\cite{ghassemi2016feasibility}, DUCK\cite{pratt2016dynamic} that use similar concept of bipedal locomotion with thrusters.

A common architecture for multi-modal systems seen today and widely considered are wheeled rotorcraft, flying cars. These transform between wheeled and aerial mobility. Many drones are fitted with active and/or passive wheels that can be used when flying is not possible. For instance, \cite{morton2017small} and \cite{kalantari2020drivocopter} classify and explain different configuration of legged and wheeled aerial systems along with their advantages and downfalls. The study also proposes a multi-modal aerial vehicle called DrivoCopter, featuring active meshed spherical wheels that surround the propellers and protect the propellers during collision. Another example is FCSTAR by \cite{david2021design}. FCSTAR is a re-configurable multi-rotor that showcases multi-modal locomotion. It has the ability to fly using propellers and also repurposes the propellers to assist in climbing nearly verticals by providing a reverse thrust. 

From these example it can be seen that multi-modal, terrestrial-aerial robots are unexplored compared to ground or aerial vehicles. particularly, multi-modal systems that aim to integrate legged and aerial locomotion due to challenging design are overlooked. The main objective of this paper is to discuss the challenges in achieving multi-modal locomotion, and to report our progress in developing our quadrupedal robot capable of multi-modal locomotion (grounded and aerial locomotion) \cite{dangol_performance_2020, dangol_thruster-assisted_2020, dangol_towards_2020, dangol2020feedback, dangol2021hzd, ramezani_generative_2021, sihite2021unilateral, liang2021rough}.

This paper is organized as follows: Section~\ref{sec:objectives} outlines our objectives and goals in developing Husky, followed by Sections \ref{sec:multimodal} and \ref{sec:path_planning} which describe the challenges in multi-modal locomotion and our progress in exploring the multi-modal locomotion aspect of our robot through simulations, Section \ref{sec:hardware} discusses the Husky's mechanical design and electrical components, Section \ref{sec:experiment} outlines the experimental set up and results of the trotting in place experiments, and finally Section \ref{sec:conclusions} for the concluding remarks.

\section{Objectives of Husky Carbon Project}
\label{sec:objectives}

\begin{figure}[t!]
    \centering
    \vspace{0.1in}
    \includegraphics[width=1.0\linewidth]{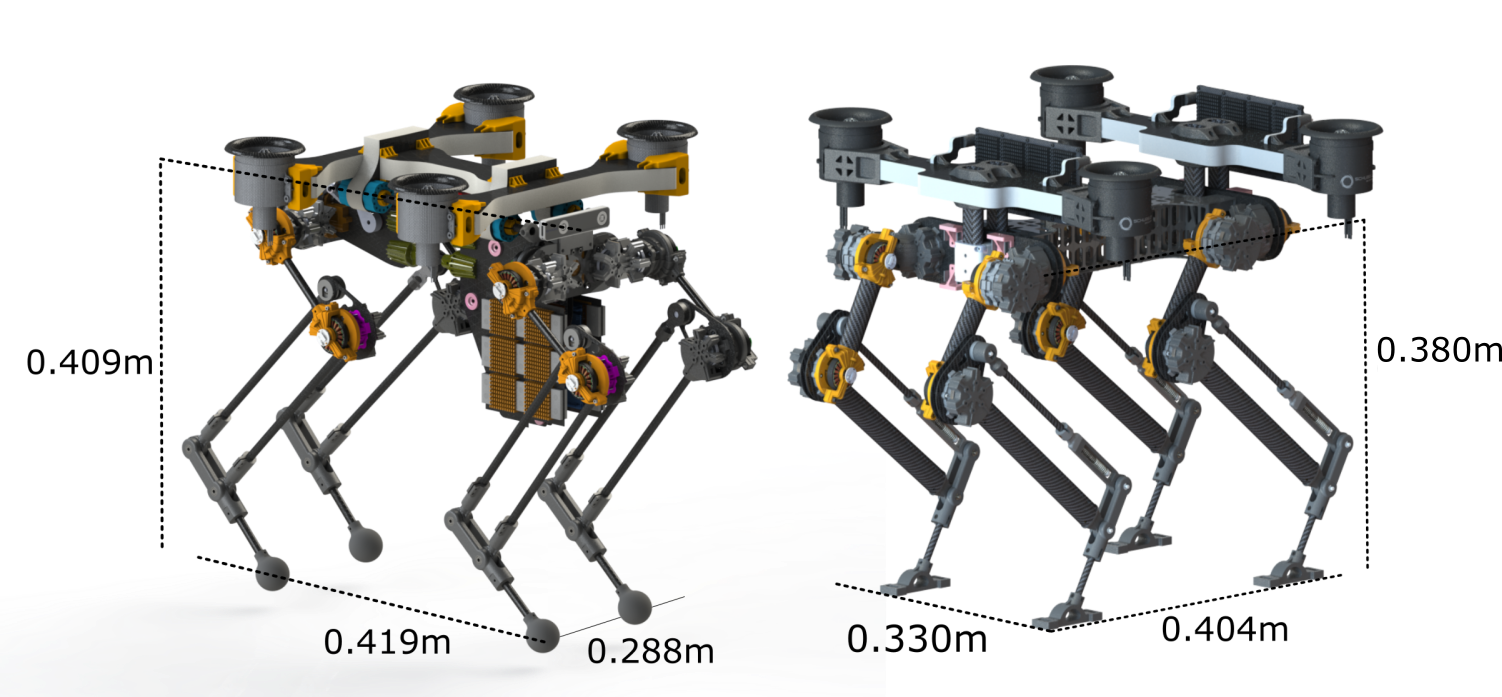}
    \caption{The CAD design of Husky v1.0 (left) and v2.0 (right) for size comparison with state-of-the-art quadrupedal robots.}
    \vspace{-0.1in}
    \label{fig:size-comparsion}
\end{figure}

The major motivation is to create a platform that possesses the fast mobility (at high altitudes) of an aerial system and the safe, agile and efficient mobility of a quadrupedal system in unstructured spaces. In Search And Rescue (SAR) operations and in the aftermath of unique incidents such as flooding, one event may accompany another disaster. A hurricane may produce flooding as well as wind damage, or a landslide may dam a river and create a flood. In these scenarios, mono-modal mobile systems can easily fail. For instance, Unmanned Aerial Systems (UAS) can deliver important strategic situational awareness involving aerial survey and reconnaissance, which can be key in locating victims quickly, through scans of the area with their suite of sensors. However, airborne structural inspection of buildings in harsh atmospheric conditions is challenging if not impossible. In addition, aerial mobility is not feasible inside collapsed buildings. To inspect inside these structures, legged mobility in form of crawling and walking is superior to aerial or wheeled mobilities \cite{alexander_principles_2013}.

Another motivation is the safety of mobility. The absence of safe UAS was costly in the aftermath of Hurricane Katrina in 2005. In this incident, which set the stage for drone deployments, regulations promulgated by the Federal Aviation Administration (FAA) posed severe limitations on drone operations in inflicted regions. According to the FAA, extreme care must be given to and when flying near people because operators tend to lose perception of depth and may get far too close to objects and people.

While energetic efficiency of legged locomotion has been extensively studied based on shaping joint trajectory \cite{ramezani_performance_2014, dangol2020towards,dangol2020performance,liang2021rough,de2020thruster}, actuator design \cite{wensing_proprioceptive_2017,seok_design_2015} and compliance \cite{kashiri_overview_2018,hutter2012starleth}, there has been little to no attempt to connect it to robot morphology. For instance, efficiency was a key factor in the MIT Cheetah robot's trotting gaits with jerky and interrupted joint movements under large Ground Reaction Forces (GRFs) \cite{park2017high,hyun_high_2014}. Efficiency was achieved based on electric actuator design and selection of predefined trotting gaits. Since commercial actuators with high ratio gearboxes are often tightly designed based on more continuous and controlled behaviors, the Cheetah possessed backdrivable and brushless actuators. 
In ANYmal, the emphasis has been on endurance and with its bulky structure, so natural and fast running patterns showcased by MIT's Cheetah are not feasible on this platform \cite{hutter_anymal_2016}. ANYmal's design characteristics are high torque joint actuation through the combination of electric motors and harmonic drives with rotational springs, precision joint movement measurement, and a rigid, bulky leg structure that accommodate high mechanical bandwidth for high bandwidth impedance control. With these characteristics, robot morphology and payload have not been important in the design of ETH's ANYmal. 

\section{Multi-modal Locomotion Challenges}
\label{sec:multimodal}

\begin{figure*}[t]
    \centering
    \vspace{0.1in}
    \includegraphics[width = \linewidth]{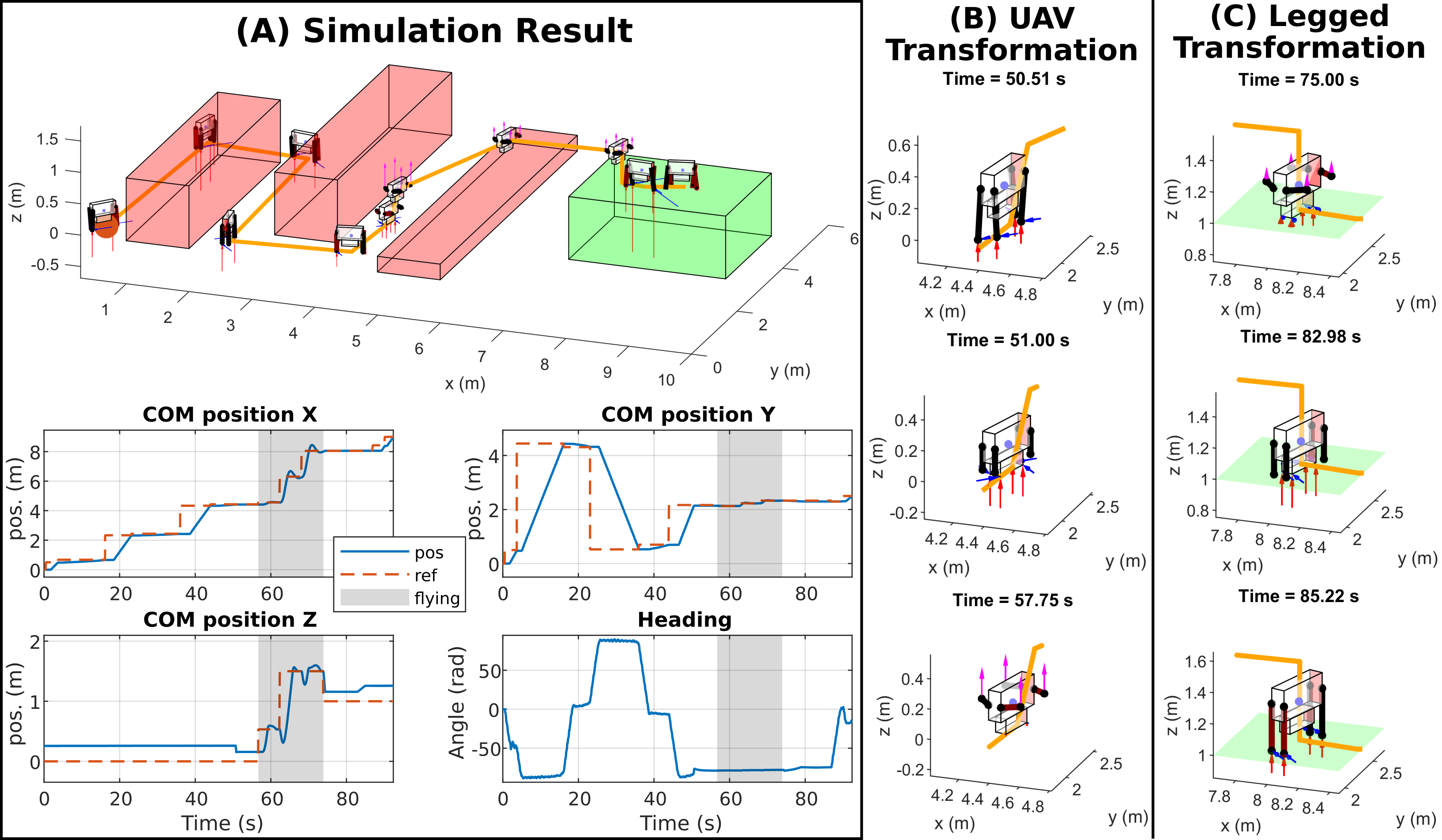}
    \caption{The simulation result for the trajectory following algorithm showing the legged and aerial mobility capabilities of Husky. In this simulation, the propellers were attached near the leg end of the robot. \textbf{(A)} Shows the trajectory followed by the robot, position states, and heading in the simulation. \textbf{(B)} Shows the transformation sequence from legged to aerial mobility. \textbf{(C)} Shows the transformation sequence from aerial to legged mobility.}
    \vspace{-0.1in}
    \label{fig:multimodal_pathplanning}
\end{figure*}

\begin{figure*}[t]
    \centering
    \vspace{0.1in}
    \includegraphics[width=0.9\linewidth]{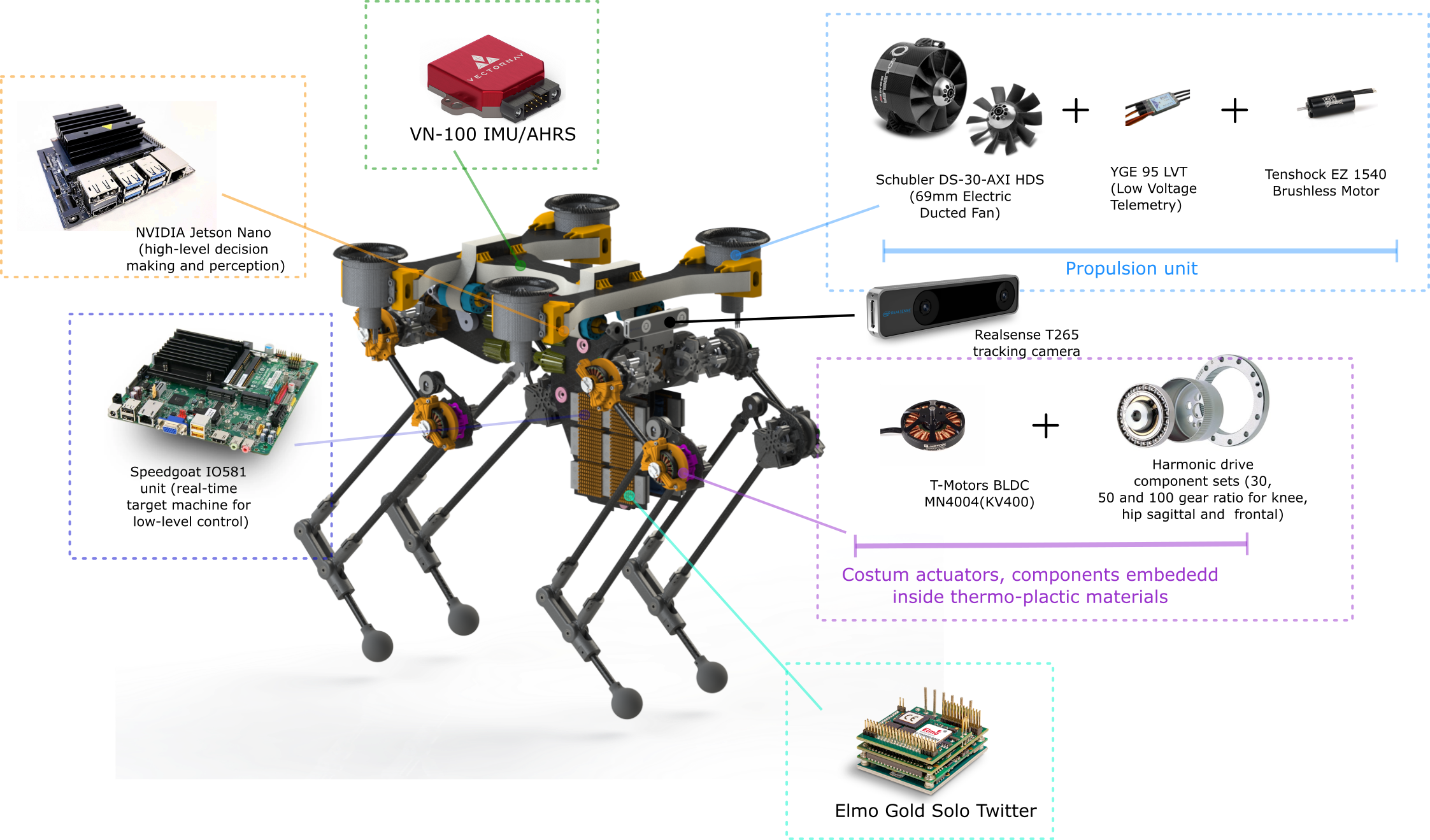}
    \caption{Overview of Husky's electronics and mechanical system. The core computing and signal processing are done by the Speedgoat IO581 which acts as the target machine and running Simulink Realtime. The motors are custom-made T-Motors MN4004 400KV, paired with Harmonic Drive of varying gear ratios using embedded additive manufacturing technology to reduce weight and the number of  fasteners in the design. The motors are driven by the Elmo Gold Solo Twitter which can drive the motor at a very high bandwidth and precision. The on-board sensors are the VectorNav VN-100 IMU, Intel Realsense T265, and Hall-effect encoders on each motor. Finally, the flight system uses DS-30-AXI HDS electric ducted fans with the appropriate brushless motor and ESC combination. }
    \vspace{-0.1in}
    \label{fig:embedding}
\end{figure*}


Compliance can yield control design challenges both for aerial and legged mobility. In general, whole body control in legged systems is challenging by itself and unplanned compliance can lead to extra  \cite{dario_bellicoso_perception-less_2016,farshidian_robust_2017}. That said, compliance is not a negative property by itself and is the defining characteristics of biological locomotion systems \cite{alexander_principles_2013}. This is the reason legged community has adopted series elastic actuators \cite{roy_hybrid_2013,hutter_efficient_2013}. 

The inherent compliance in Husky's body has led to a few issues which have been addressed with the help of closed-loop feedback. While Husky's compliant legs can potentially accommodate locomotion on grounds that feature unknown irregularities, body flexibility in Husky can prohibit precise kinematic planning for perceived ground changes and can lead to instantaneous body sagging each time feet touch-downs occur. Other issues including degraded mechanical bandwidth as it is important to the projection of joint torques to body forces, impedance control, or foot placement has been observed in Husky. For instance, the leg compliance can introduce oscillations of large amplitude and can lower the natural frequency of the tracking controller meant to position the COM at desired locations with respect to the contact points \cite{townsend_mechanical_1989,fankhauser_robust_2018,hutter_efficient_2013,winkler_gait_2018,dario_bellicoso_perception-less_2016}.


Quadrupedal robots are often designed in such a way that they have wide support polygons which allow trivial, quasi-static locomotion \cite{winkler_gait_2018,mastalli_-line_2015,farshidian_robust_2017,farshidian_robust_2017}. Husky possesses a very small stability margin based on a crude definition \cite{kimura_adaptive_2007} which considers the shortest distance from the robot's projected COM to the edges of its support polygon constructed by the contact points. This small margin is led because of two reasons. First, the small stability margin is caused partly because of a low cross-section torso -- i.e., the hips are located very close to each other in the frontal plane -- which is important to reduce induced drag forces for aerial mobility. Second, it is partly caused by the inherent compliance in the system which can cause fall-overs even when the robot has four contact points. Due to the robot body flexibility COM can shift towards the edges of the support polygon very easily if required corrections are not made quickly through foot placement or active control of the joints.     

Small stability margins in Husky has led to other challenges such as gait design issues that generally are expected in bipedal systems and not quadrupedal robots \cite{apgar_fast_2018}. In Husky, the stability margin and gait cycle periods are adversely related. When the gait cycle time is larger than approximately one third of a second, which is equivalent to a 3-Hz gait cycle, even when the robot retains three contact points with the ground surface the projected position of the robot COM could reach to the boundaries of the support polygon. If new foot placements are not involved the robot's stability margin will reduce further by the sagging effects led by the compliance in the robot. This has posed severe gait design challenges for Husky.

\section{Progress on Path Planning}
\label{sec:path_planning}

\begin{figure}[t]
    \centering
    \vspace{0.1in}
    \includegraphics[width=\linewidth]{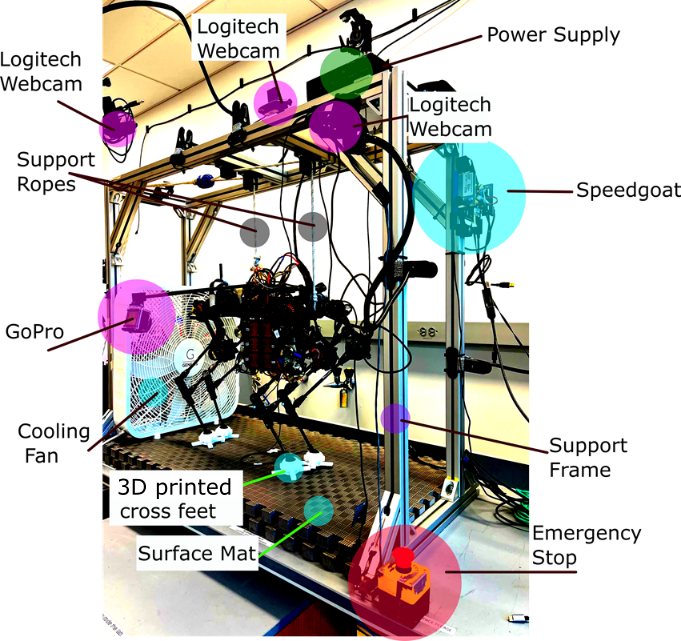}
    \caption{Husky test arena. This arena is equipped with multiple cameras at several angles, a Speedgoat which acts as the target computer, a textured surface mat, power supply for powering the robot, and a large fan to help cooling down the actuators. During the experiment, an AprilTag is attached to the support frames in front of the robot to be tracked by the Realsense camera.}
    \vspace{-0.1in}
    \label{fig:arena}
\end{figure}

\begin{figure*}[t]
    \centering
    \vspace{0.1in}
    \includegraphics[width=\linewidth]{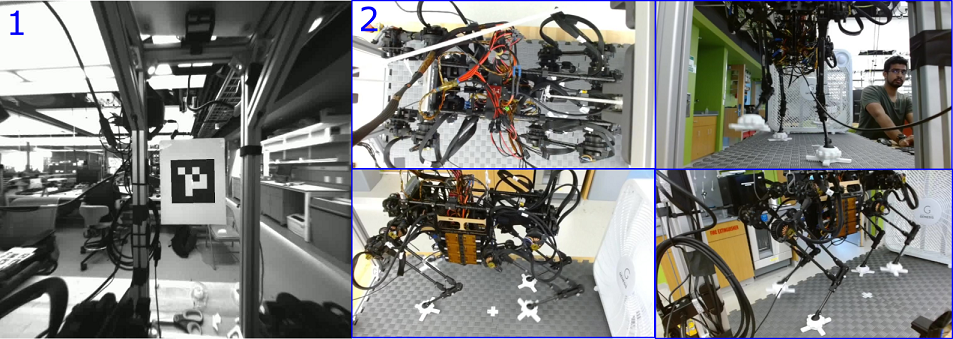}
    \caption{Snapshots of the Husky trotting inside the test arena. The arena is equipped with cameras at various angles to capture the experiment in great detail. An AprilTag is placed in front of the robot which is tracked by the Realsense camera which is used to maintain a stable pose. }
    \vspace{-0.1in}
    \label{fig:trotting}
\end{figure*}

In order to take full advantage of the multi-modal capacities of the Husky, it is necessary to develop a path planning optimization methods that can incorporate Husky's multi-modal locomotion capability. Numerous researches have already been done on multi-modal robots which are either able to roll and fly such that the HyFDR \cite{sharif_energy_2018}\cite{sharif2019new} and the Drivocopter \cite{suh_optimal_2019} or to drive and navigate on water such that the Ambot \cite{suh_optimal_2019}. Most of the methods developed in these articles use a uniform discretization of the space, and then the optimal path is found with the Dijkstra algorithm \cite{suh_optimal_2019}, or with the A$^{\star}$ \cite{sharif_energy_2018}\cite{araki_multi-robot_2017}. 

Furthermore, in \cite{suh_optimal_2019}, an optimization technique based on a reduced model of the system is used to calculate the costs of the edges and then to smoother the final trajectory. Araki et al. \cite{araki_multi-robot_2017} have coupled their path planning method to a prioritization algorithm allowing swarm operation with 20 flying cars. While in this article \cite{sharif2019new}, Sharif et al. have developed an algorithm to select the locomotion mode of the HyFDR robot allowing to optimize the transport cost during outdoor navigation with only a 2D map of the environment. 

We developed a simulator using a reduced-order model (ROM) to simplify the trajectory tracking and cost calculations in the path search algorithm. This ROM assumes massless leg linkages and can be reduced down to a single body, 6 DOF dynamics. In this simplified model, each leg has 3 DOF to describe the foot position. These 3 DOF of leg $i = {1,2,3,4}$ are the hip frontal angle ($\phi_i$), hip sagittal angles ($\psi_i$), and leg length ($l_i$). This results in a total of 12 kinematics DOF and 6 dynamical DOF which is much simpler than the full dynamical model of the robot. The dynamical model can be derived using Euler-Lagrangian formulation. 

Here, we assume a conventional flight control design which is skipped for brevity of this report. However, the optimality of the low-level legged locomotion control in terms of achieving feasible gaits is enforced within an RG-based framework. The RG framework is utilized to enforce the friction pyramid constraint by manipulating the applied reference into the kinematic states \cite{dangol2020performance, liang2021rough,sihite_integrated_2021}. This method is very useful as it avoids using optimization frameworks to enforce locomotion feasibility constraints which as a result facilitates faster high-level decision making.

Two different discretization methods have been used to create a set of nodes and edges representing the environment, and their performances are then compared. The first one consists in dividing the space into a set of uniformly distributed points. While, in the second one, the 3D environment is discretized into a set of nodes and edges with the 3D MM-PRM. Like in \cite{MM_PRM}, this adapted version of the Probabilistic Road Map (PRM) algorithm takes into account the Multi-Modal nature of the robot's movements. The simulation results of the robot navigating the environment using the path generated by the path-planning algorithm can be seen in Fig.~\ref{fig:multimodal_pathplanning}. The details of the path planning algorithms and the simulations will be reported in a separate work. 

\section{Brief Overview of Husky Design}
\label{sec:hardware}

In this section, we briefly provide an overview on the mechanical design and on-board electronics, including computing unit, proprioceptive and extroceptive sensors employed in the design of Husky.  

\subsection{Mechanical Design}

Husky, shown in Fig.~\ref{fig:cover-image}, when standing as a quadrupedal robot, is around 0.4 m tall. The robot is about 0.3 m wide. It is fabricated from reinforced thermoplastic materials through additive manufacturing and weighs 4.3 kg. It hosts on-board power electronics and, currently, it operates using an external power supply. The current prototype has a stereo tracking camera as the exteroceptive sensor, but lacks LiDAR which we consider adding in our future protoypes. 
The robot is constructed of two pairs of identical legs in the form of parallelogram mechanisms. Each with three degrees-of-freedom (DOFs), the legs are fixated to Husky's torso by a one-DOF revolute joint with a large range of motion. In the current iteration, we designed a mount for four thrusters on top of the robot with a configuration typically seen in quadcopters for flying. 

The robot possesses a total of twelve joint actuators. Each custom-made actuator has a pancake brushless-DC motor winding from T-motor with a KV equal to 400. Harmonic drive component sets (flexspline, circular spline and wave generator) with gear ratios: 30, 50 and 100 for knee, hip sagittal and hip frontal, respectively, and hall-effect-based incremental encoders are embedded inside the 3D-printed structure of the robot. The embedding process has minimized the use of metal housings and fasteners and has yielded a lighter structure. The current integration has made Husky self-sustained for stable, walking and trotting gaits.

\subsection{Electronics}



Proprioceptive state estimators utilising IMU and robot's kinematics~\cite{8118116, doi:10.1177/0278364919894385} proved to show low odometry drift~\cite{8460731} that elevation mapping can be used for statically stable walking (e.g. when friction is high and speed is low). 
However, for dynamic locomotion such as trotting\--- when slippage is likely to occur\--- these estimators cannot precisely predict the elevation in front of the robot so that the robot can execute safe step planning. Therefore, we aim to integrate proprioceptive measurements with an exteroceptive sensor such as a camera as the integration of cameras with IMU measurements have demonstrated low odometry drift~\cite{7557075, mohta2018fast} suitable for quadruped robot's motion planning~\cite{8790726}. 
The current iteration of Husky utilizes an on-board IMU (VectorNav VN-100) and tracking camera (Intel Realsense T265) to help it balance itself during walking and trotting. These are interfaced with a Speedgoat IO581 Real Time Target Machine that is running an instance of the robot's Simulink model. The Speedgoat unit interfaces with Elmo Servo Drives through an EtherCAT connection to control the motors.



\section{Tethered Trotting Tests}
\label{sec:experiment}

\begin{figure}
    \centering
    \vspace{0.1in}
    \includegraphics[width=\linewidth]{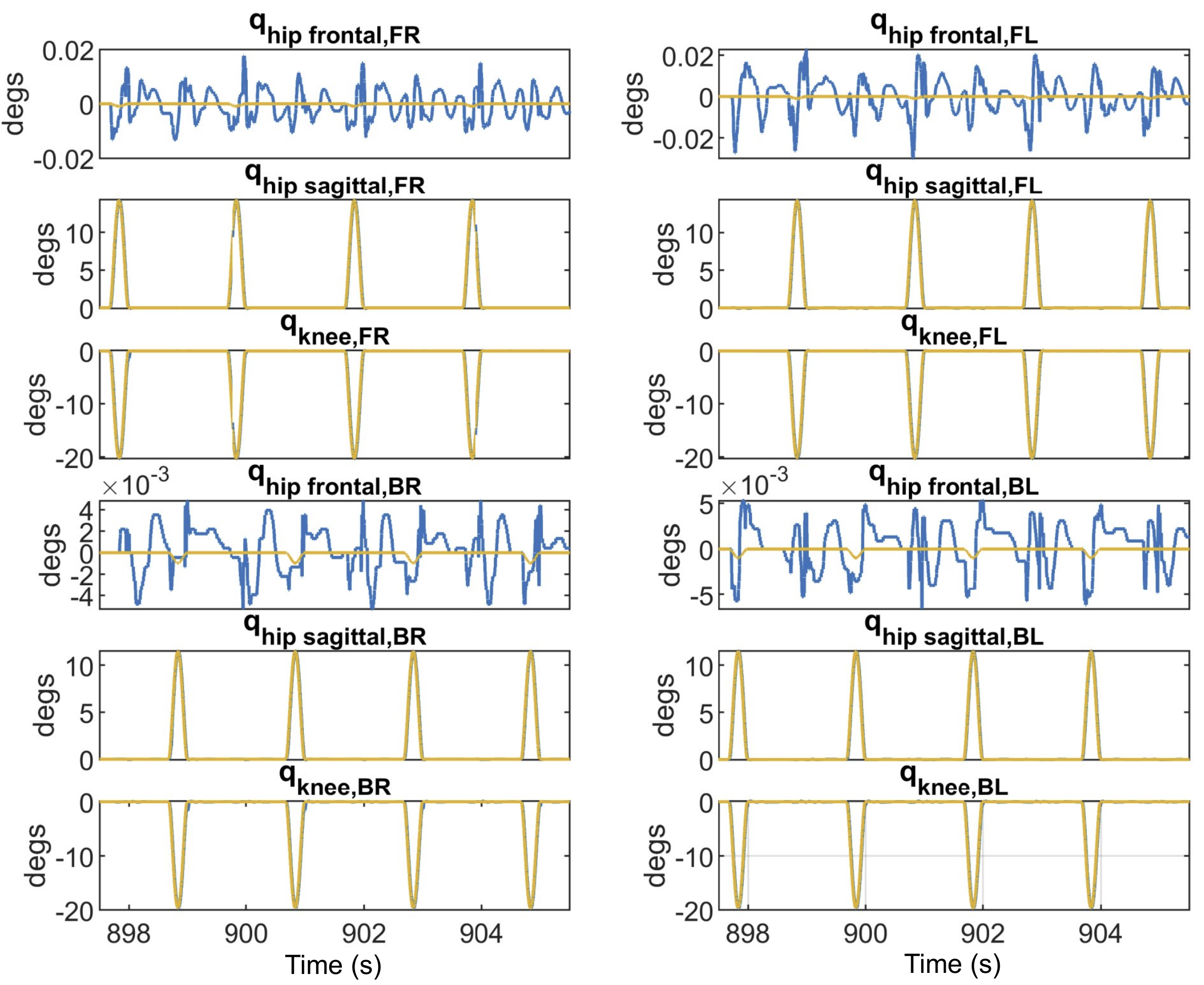}
    \caption{Time-evolution of joint trajectories during the experiment. Yellow lines are the target joint angle, while the blue line are the joint states as measured by the encoders.}
    \vspace{-0.1in}
    \label{fig:joint-traj}
\end{figure}

\begin{figure}
    \centering
    \vspace{0.1in}
    \includegraphics[width=\linewidth]{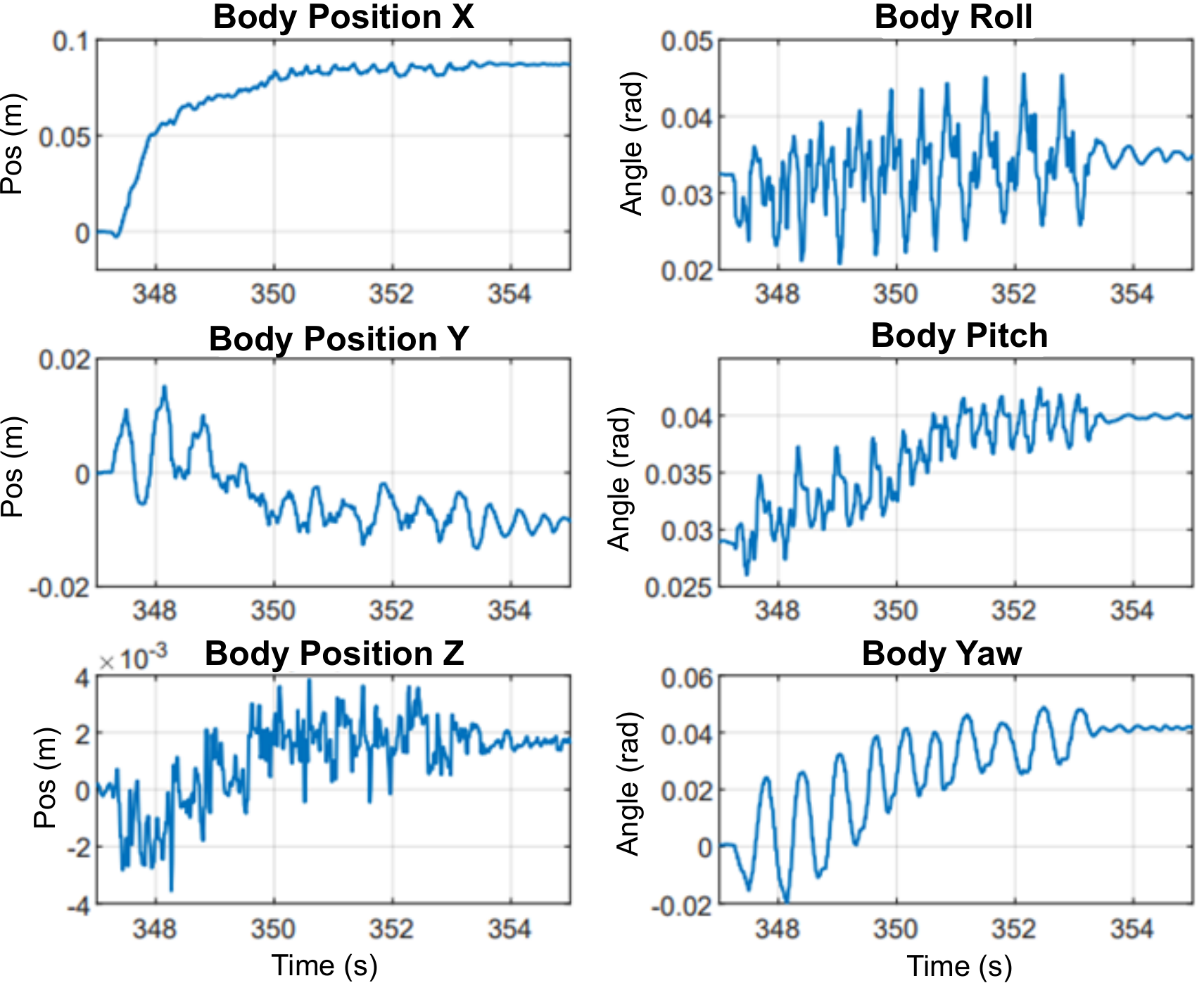}
    \caption{Plots of Husky's position and orientation during the experiment. There are some shifts in the pose from its initial position into a stable limit cycle.}
    \vspace{-0.1in}
    \label{fig:plot_pose}
\end{figure}

\begin{figure}
    \centering
    \vspace{0.1in}
    \includegraphics[width=\linewidth]{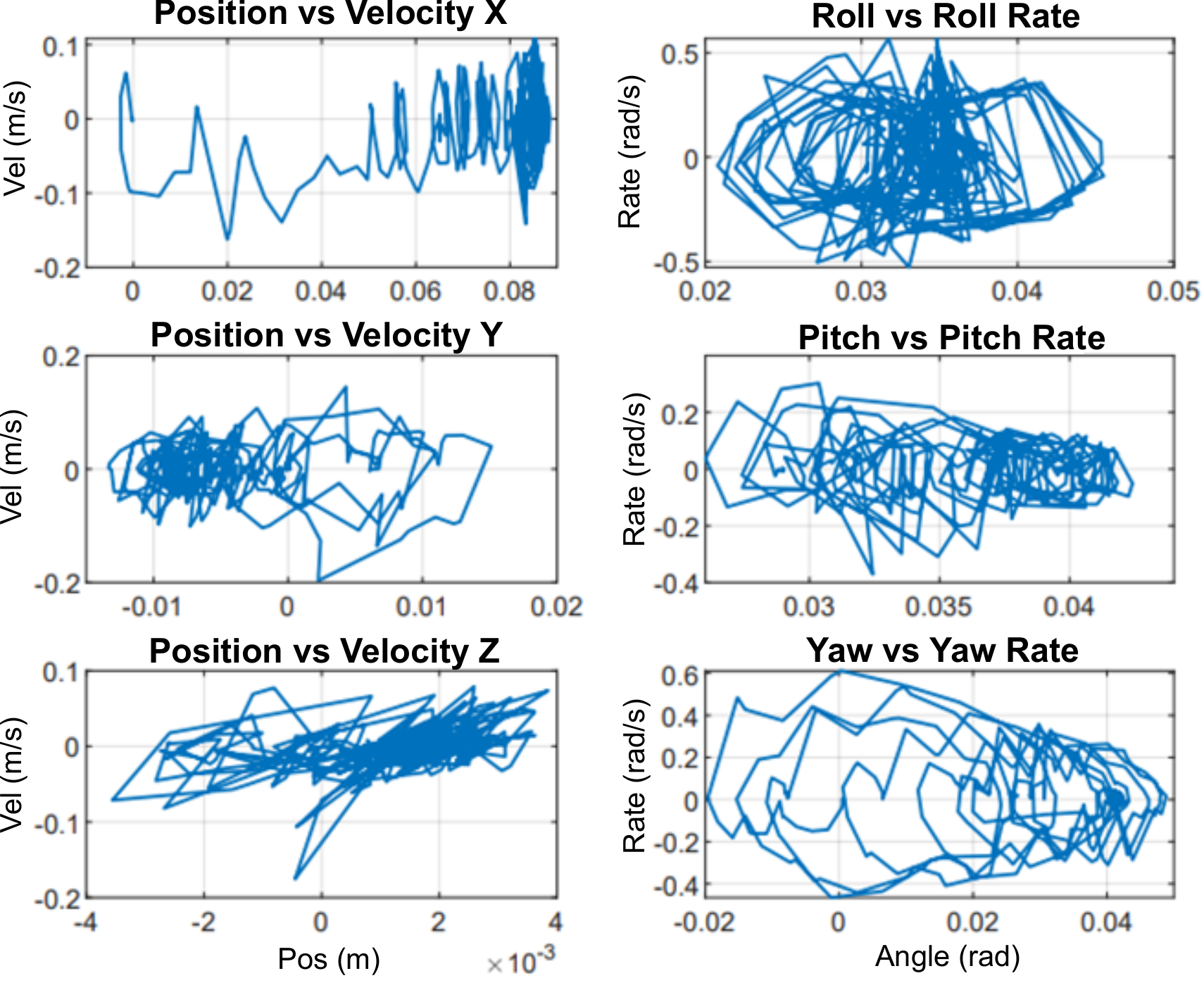}
    \caption{Phase plots of Husky's Poses and their rates during the experiment. The phase plot shows that the trotting gait is stable and reached a stable limit cycle.}
    \vspace{-0.1in}
    \label{fig:plot_phase}
\end{figure}

We tested the performance of Husky in our test arena, shown in Fig. \ref{fig:arena}. The arena is equipped with cameras for observing gaits from different angles, a padded ground surface mat, an external power supply and a cooling fan to mitigate occasional motor heat-up during testing. We utilize this arena to test the Husky's controller performance as the robot walks in place tethered to the external power supply. A pulley mechanism attached to the support ropes allows us to control the amount of weight carried by Husky's legs, allowing for safe testing and simulation of gaits under partial propulsion from the thrusters. 

To stabilize the robot as it is walking, an AprilTag is placed in front of the robot visible to an Intel RealSense onboard camera, providing pose information to the controller as feedback. 
Using this setup, we have been able to demonstrate three contact and two contact trotting in place at various gait speeds.

Figure \ref{fig:trotting} shows the set up and snapshots of Husky's tethered trotting experiment, and experimental data can be seen in Figures \ref{fig:joint-traj} to \ref{fig:plot_phase}. Figure.~\ref{fig:joint-traj} shows the joint trajectories of the robot for a four loops of a two contact trotting gait. The graphs show the trajectories for each of the three joints in the four legs, Front Right (FR), Front Left (FL), Back Right (BR) and Back Left (BL). The graph shows that desired trajectories are tracked very well by the controller. Fig.~\ref{fig:plot_pose} shows the pose of the robot (position and orientation) across the movement. The body remains fairly stable although noise is introduced due to inherent compliance in the joints. Fig.~\ref{fig:plot_phase} shows the phase portrait of the robot during the trotting experiment depicting Husky's convergence to a stable limit cycle.

\section{Conclusions and Future Work}
\label{sec:conclusions}

In this work, we present the motivation, objectives, and progress of our multi-modal quadrupedal robot, the Husky Carbon. We provided detailed information on the practical challenges of developing this robot and present our most recent robot design along with the simulation and experimental results. For future work, we look into expanding the capabilities of the platform by installing a Lidar sensor for accurate localization and mapping including elevation mapping which is useful for navigating unstructured environment. As discussed earlier, inherent compliance in Husky caused by it's unique lightweight design has led to issues that have been partially addressed. 
We shall fully address this issue by predicting the compliance and body's center of mass using a model-based approach for a more accurate kinematics and dynamics stability.










\IEEEtriggeratref{47} 
\printbibliography

\end{document}